%% file: acl2023.tex
\pdfoutput=1

\documentclass[11pt]{article}

\usepackage{ACL2023}

\usepackage{times}
\usepackage{latexsym}

\usepackage[T1]{fontenc}

\usepackage[utf8]{inputenc}

\usepackage{microtype}

\usepackage{inconsolata}

\usepackage{graphicx}
\usepackage{booktabs}
\usepackage{todonotes}
\usepackage{multirow}
\usepackage{rotating}
\usepackage{amsmath}
\title{Learning Multimodal Cues of Children’s Uncertainty 
}

\author{
    Qi Cheng\textsuperscript{\rm 1}\textsuperscript{*},
    Mert İnan\textsuperscript{\rm 4}\textsuperscript{*},  
    Rahma Mbarki\textsuperscript{\rm 3}\textsuperscript{*}, 
    Grace Grmek\textsuperscript{\rm 2}\textsuperscript{*}, 
    Theresa Choi\textsuperscript{\rm 1}\textsuperscript{*}, \\
    \textbf{Yiming Sun}\textsuperscript{\rm 3},
    \textbf{Kimele Persaud}\textsuperscript{\rm 3},
    \textbf{Jenny Wang}\textsuperscript{\rm 3},
    \textbf{Malihe Alikhani}\textsuperscript{\rm 4} \\
    \textsuperscript{\rm 1} University of Pittsburgh, PA, USA \ \ 
    \textsuperscript{\rm 2} Harvard Medical School, MA, USA \\
    \textsuperscript{\rm 3} Rutgers University, NJ, USA \ \
    \textsuperscript{\rm 4} Northeastearn University, MA, USA \\
    \texttt{\{qic69, tec63\}@pitt.edu,} \texttt{ggrmek@mgh.harvard.edu,} \\ \texttt{\{rm1218, kjg117, jinjing.jenny.wang\}@rutgers.edu,} \\ \texttt{\{inan.m, alikhani.m\}@northeastern.edu}
}

\begin{document}
\maketitle
\begin{abstract}
Understanding uncertainty plays a critical role in achieving common ground \cite{clark1983common}. This is especially important for multimodal AI systems that collaborate with users to solve a problem or guide the user through a challenging concept. 
In this work, for the first time, we present a dataset annotated in collaboration with developmental and cognitive psychologists for the purpose of studying nonverbal cues of uncertainty. We then present an analysis of the data, studying different roles of uncertainty and its relationship with task difficulty and performance. Lastly, we present a multimodal machine learning model that can predict uncertainty given a real-time video clip of a participant, which we find improves upon a baseline multimodal transformer model. This work informs research on cognitive coordination between human-human and human-AI and has broad implications for gesture understanding and generation. 
The anonymized version of our data and code will be publicly available upon the completion of the required consent forms and data sheets. 

\end{abstract}

\input{sections/0_introduction}
\input{sections/1_related-work}
\input{sections/2_data}
\input{sections/3_analysis}
\input{sections/4_computational}
\input{sections/5_results}

\input{sections/6_conclusion}

\bibliography{refs}
\bibliographystyle{acl_natbib}
\appendix

\input{sections/appendix}

\end{document}

%% file: sections/0_introduction.tex
\section{Introduction}





Recognizing uncertainty in interlocutors plays a crucial role in successful face-to-face communication, and it is critical to achieving common ground \cite{clark1983common}. To accurately identify uncertainty signals, human listeners learn to rely on facial expressions, hand gestures, prosody, or silence.  
AI systems that aim to collaborate and coordinate with users in a human-like manner also need to understand these signs of uncertainty. To this end, in this paper, we introduce a multimodal, annotated dataset for uncertainty detection in young children. 




\begin{figure}[t]
    \centering
    \includegraphics[width=\columnwidth]{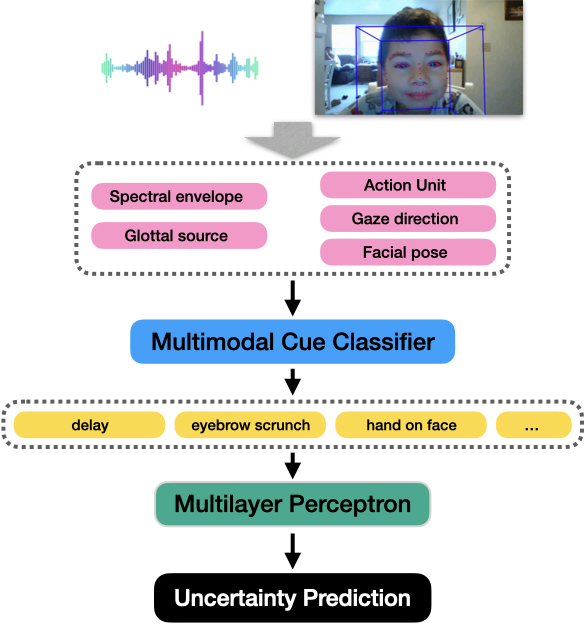}
    \caption{A diagram of our multimodal machine learning model. After identifying uncertainty cues in the multimodal transformer, the model passes the cues onto a final multilayer perceptron classifier to output whether the child is expressing uncertainty or not.}
    \label{fig:my_label}
\end{figure}


As a multimodal communicative sign, identifying uncertainty is an important and challenging task for AI systems. Especially because it varies across different ages and demographics; it is sometimes verbalized and sometimes not \cite{blanco}; it brings in different modalities, and it is subtle. Although it is critical, uncertainty signal recognition is understudied in younger children. In this work, we study detecting uncertainty in the setting of a counting game for children ages 4-5. We first identify potential cues of uncertainty presented in different modalities (e.g., spontaneous verbal responses, hand gestures, facial expressions, hesitation) and specifically examine the relationship between task difficulty, task performance, and exhibited levels of uncertainty. We then use these cues to inform an ensemble model, which first identifies these cues from multimodal data and then uses them to predict uncertainty (Figure \ref{fig:my_label}).

This work informs research on cognitive coordination between human-human and human-AI collaboration. With this paper, we contribute an annotated multimodal dataset of uncertainty in children (Section \ref{sec:data} provides details about the dataset, annotation protocol, and analyses of the dataset); we analyze the performance of multimodal transformer models in identifying uncertainty on this dataset (Sections \ref{sec:computational} and \ref{sec:results}); and finally we present a case study on how children express uncertainty based on their age.




%% file: sections/1_related-work.tex
\section{Related Work}
\label{sec:related-work}
We cover related works of uncertainty in two sections: datasets and protocols for studying uncertainty in children, the Approximate Number System, and uncertainty in human-AI interactions.

\subsection{Datasets and Protocols for Studying Uncertainty in Children}
Adults are generally more direct and communicate their uncertainty via explicit verbal cues. Children, however, lack this insight into their own uncertainty, making uncertainty detection more difficult from an outsider’s perspective. As such, detecting uncertainty in children remains a complex problem. 

What has been established, however, is that children consistently communicate their uncertainty through the use of various facial, auditory, and gestural cues. For example, \citet{harris} found that children are very expressive when they are uncertain. In the presence of an adult, these expressions may be communicated via hand flips, questions, and utterances, such as “I don’t know.” However, when children are alone, these same signals can be representative of signals of uncertainty. In the past, researchers have attempted to codify behaviors associated with communicating uncertainty by parsing through these various cues and creating annotation protocols. 

Previously, researchers \citet{swerts2005audiovisual} aimed to detect uncertainty in audiovisual speech by coding for different audiovisual cues in both adults and children. Their protocol consisted of audio cues (e.g. speech fillers and speech delays) and facial movements (e.g. eyebrow movement and smiling). While the protocol included both audio and visual cues, the cues that were noted were limited. Another protocol developed by \citet{mori2019look} studied solely visual cues signaling uncertainty in speech communication. These cues included changes in gaze direction, facial expressions, and embarrassed expressions. An additional protocol developed by \citet{bonfig} studied uncertainty through facial expression entirely. 

However, while these protocols are indeed useful, they lack the specificity necessary for our goal of pinpointing various multimodal cues associated with uncertainty. There are other various protocols, but they are also limited, typically adhering to one modality. Consequently, we expanded upon these existing protocols and included other multimodal behaviors grounded in developmental and cognitive psychology, and presented multimodal machine learning models that can predict these cues and their association with uncertainty. 

Children have an intuitive sense of numbers relying on the Approximate Number System (ANS). The ANS obeys Weber’s Law, where one’s ability to differentiate between two quantities depends on the ratios of those quantities \citep{dehaene_2011, odic}. The smaller the ratio, the more difficult it is to discriminate between quantities and the more uncertainty there is in the participants’ internal representations. Previous research showed that children perform better on a numerical comparison task when given a scaffolded, Easy-First numerical task starting with easier trials (e.g., 10 vs. 5) and progressing to harder ones (e.g., 10 vs. 9), compared to children seeing the same exact trials in the reversed order (i.e., Hard-First), an effect termed “confidence hysteresis” \cite{odic2014hysteresis}. This implies that confidence is built by gradually working up to harder tasks, resulting in better performance, whereas starting out with more difficult tasks reduces confidence, resulting in worse performance.

Due to its effectiveness at generating confidence or lack thereof in participants, such a numerical comparison task would be the ideal method for measuring behaviors associated with uncertainty. As such, the present study aims to fulfill this objective by implementing this “confidence hysteresis” paradigm into the task children are given.

\subsection{Studying Uncertainty in Human-AI Interaction}
Multimodal models have been shown to improve performance on certain tasks by grounding some aspects of the human condition with features beyond text. 
Leveraging multiple modalities is particularly applicable in cases where text may miss key insights, such as sarcasm detection \cite{castro-etal-2019-sarcasm}, depression prediction \cite{mm-depression-0}, sentiment detection \cite{mm-sentiment-detection}, emotion recognition \cite{morency-senti}, and persuasiveness prediction \cite{santos-etal-2016-domain}. Tasks involving such complex labels benefit from multiple modalities due to the richness of the data streams. In addition to understanding what is said, understanding how it is said (pitch, facial expression, body language, gesture) is crucial \cite{arg-for-mm}.

There have been attempts at predicting uncertainty from the audio through prosodic features. \citet{inbook} reported that prosodic features were successful in detecting speaker uncertainty in spoken dialogue with a 75\% accuracy. \citet{article} had similar findings with prosodic features, and self-reported states of certainty and perceived states have strong mismatches. In this paper, we address this by controlling task difficulty to affect a participant's level of observed uncertainty. A seminal study on the understanding and generation of multimodal uncertainty cues exists by \citet{Stone2008}. Here the authors analyze adult human-human conversations for uncertainty cues and try to replicate them using avatars. Our experimentation and modeling efforts, on the other hand, are focused on the domain of uncertainty detection in younger children.


%% file: sections/2_data.tex
\section{Data}
\label{sec:data}
\noindent \textbf{Participants} A group of 68 children between the ages of 4 and 5 years old ($M_{age}$ = 5;0; $SD_{age}$ = 6.88 months; 28 females) was recruited through Lookit, an online platform for developmental studies \cite{lookit}. Thirty-six parents identified their child as White, six as Asian, three as Hispanic, Latino, or Spanish origins, and the rest as multi-racial. All but three parents reported having a college degree or higher level of education. After completing the study, compensation was sent in the form of a \$5 gift card. Each child participated in 30 trials which are, on average, 8 seconds long. In total, our data is composed of 16,320 seconds of video data.

\begin{figure}[t]
    \centering
    \includegraphics[width=\columnwidth]{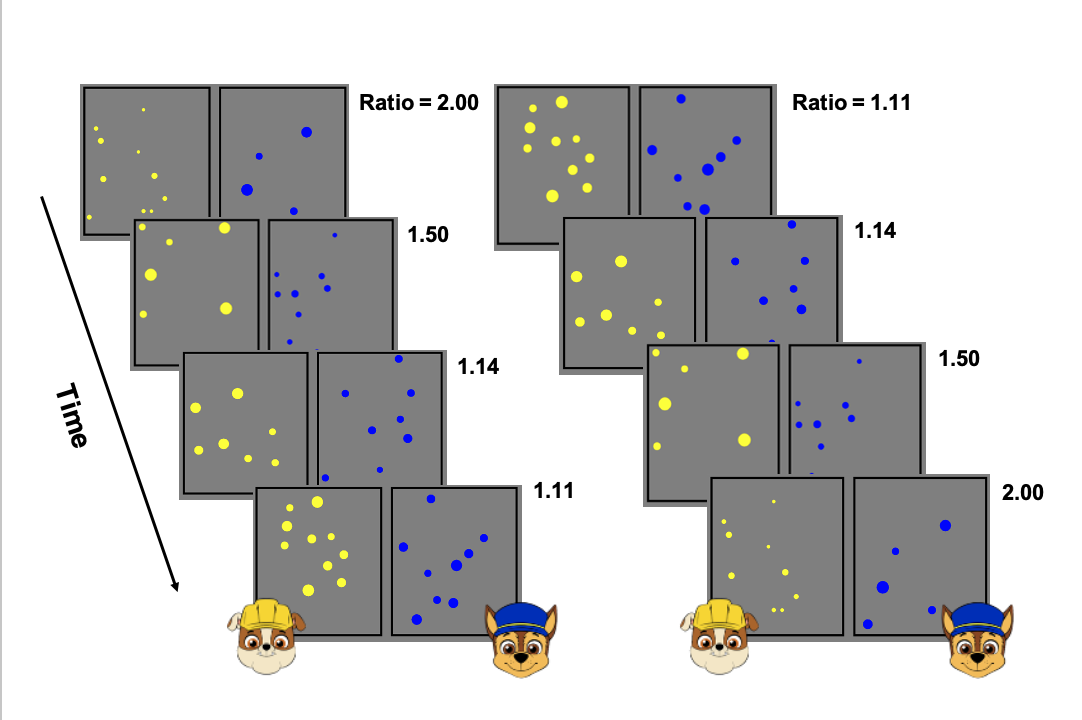}
    \caption{Schematic of experimental procedure depicting the Easy-First condition on the left and the Hard-First condition on the right. As time progresses throughout the task, the trials advance from easier ratios (2.0) to hard ratios (1.11) in the Easy-First condition. Whereas in the Hard-First condition, trials move in reverse order from hard ratios (1.11) to easy ratios (2.0) as time progresses.}
    \label{fig:task}
\end{figure}

\noindent \textbf{Task} Participants were given an Approximate Number System manipulation task adapted from \citet{Wang2021Mar} designed to impact children’s certainty about numerical quantities. Children were presented with two arrays of dots paired with two cartoon characters (Figure \ref{fig:task}) and asked to guess which character has more dots. 

Both characters and their array of dots appeared for 2500 ms before disappearing. This short display duration was chosen to ensure that children did not have sufficient time to count. Children were then asked to 
click on the side of the screen showing the greater number of dots. Children were given immediate audio feedback for each trial once they chose their response. 

Children completed 30 trials with the following number pairs: 10:9 dots (1.11 ratio), 8:7 (1.25 ratio), 14:12 (1.17 ratio), 10:8 (1.13 ratio), 9:6 (1.5 ratio), and 10:5 (2 ratio). In half of the trials, arrays with more dots had a greater, congruent cumulative area. In the other half of the trials, arrays with the greater number of dots had a smaller, incongruent cumulative area.

Children were randomly assigned to either the Easy-First or Hard-First conditions. In the Easy-First condition, trials advanced from the easier trials (e.g., 10:5) to the harder trials (e.g., 10:9) in a staircase order following the design of \citet{Wang2021Mar}. Whereas in the Hard-First condition, trials move in reverse order from hard ratios (e.g., 10:9) to easy ratios (e.g., 10:5).

\noindent \textbf{Annotation Procedure} Annotators first watched the video muted so as not to be influenced by the vocal feedback from the task since whether or not the child answered right or wrong may lead them to over/under-interpret certain cues. During this first watch, they marked all present physical cues as indicated by the protocol. On their second watch, annotators unmuted the video, and marked all verbal cues. If the cue was not present, the corresponding cell was left empty.

\subsection{Annotation Protocol}
In collaboration with a team of developmental psychologists and cognitive psychologists, we have collected and designed a protocol that aims to study uncertainty, particularly expressed non-verbally. 
Through pilot studies observing and annotating our data, we iteratively defined our protocol and constructed a list of signals we observed as signs of uncertainty.
Our protocol spans multiple modalities and includes facial, gestural, and auditory cues to account for a broad spectrum of possible behaviors.
The protocol can be found in Table \ref{tab:protocol} with supplemental example images in Figure \ref{fig:samples}. The Rutgers University Institutional Review Board approved the research, and all parents of this study's children provided verbal consent before their children’s participation. However, only some of the parents agreed to allow their children's video and voice recordings to be shared publicly.

\begin{figure}[t]
    \centering
    \includegraphics[width=1.05\columnwidth]{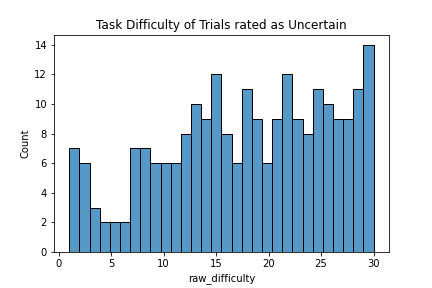}
    \caption{The distribution of uncertain trials with task difficulty on a scale of 1 (easiest) to 30 (hardest). Uncertainty shows a strong correlation with task difficulty ($r(58)=-.927, p < .01$).}
    \label{fig:distribution}
\end{figure}

%% file: sections/3_analysis.tex
\subsection{Analysis}
\label{sec:analysis}
In this section, we provide an analysis of our annotated data, identify any significant cues that contribute to detecting uncertainty, and explore when different cues occur.

\begin{table*}[h]
    \centering
    \begin{tabular}{|l|l|}
        \toprule
        \textbf{Cue} & \textbf{Description} \\
        \midrule
        \textbf{Delay}        & The participant delayed their decision-making with a pronounced pause                                     \\
        \textbf{Eyebrow raise}             & The participant raised their eyebrows                                                      \\
        \textbf{Eyebrow scrunch}           & The participant markedly scrunched their eyebrows or squinted their eyes                                              \\
        \textbf{Filled pause} & Utterances such as “umm,” “hmm,” or “uh.”\\
        \textbf{Frustrated noise}& Sounds of verbal frustration, such as sighing, groaning, and growling\\
        \textbf{Funny face}                & The participant grimaced or made an unconventional facial expression                                                          \\
        \textbf{Hand on face}              & Any kind of movement that includes a participant putting a hand on their face \\
        \textbf{Head tilt}                 & The participant tilted their head to either side while making their decision\\
        \textbf{Look away}                 & The participant was distracted and not paying attention to the task                                          \\

        \textbf{Look to adult}             & The participant looked towards their parent when making their decision\\
        
        \textbf{Shoulder movement}         & The participant made a pronounced shoulder movement, such as shrugging                                                  \\
        \textbf{Smile}                     & The participant smiled                                             \\
        
        \textbf{Verbal cues}  & Any spoken words   \\
    \bottomrule
    \end{tabular}
    \caption{Categories in our annotation protocol.} 
    \label{tab:protocol}
\end{table*}

 \begin{figure*}[t]
    \includegraphics[width=.35\columnwidth]{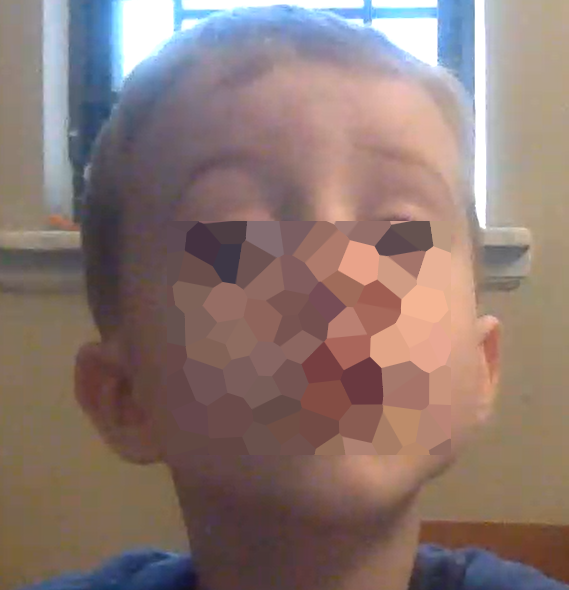}\hfill
    \includegraphics[width=.35\columnwidth]{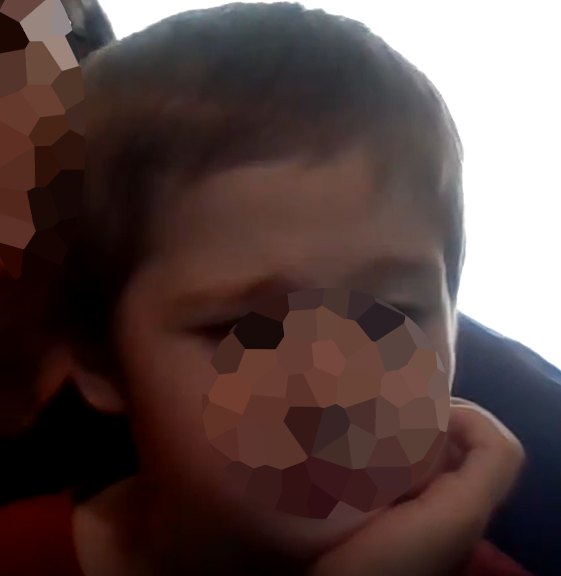}\hfill
    \includegraphics[width=.35\columnwidth]{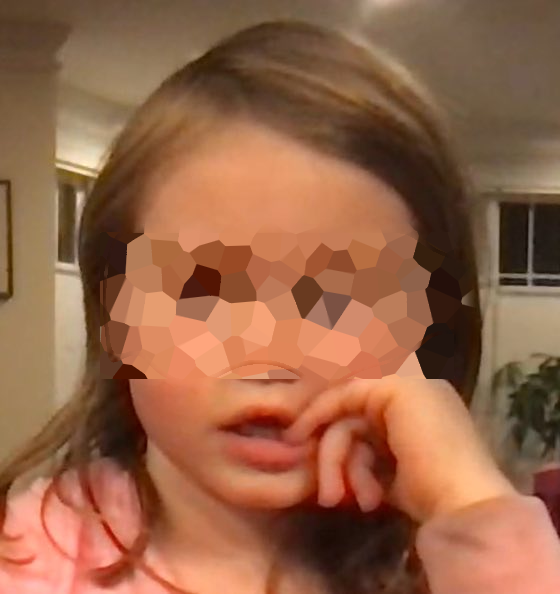}\hfill
    \includegraphics[width=.35\columnwidth]{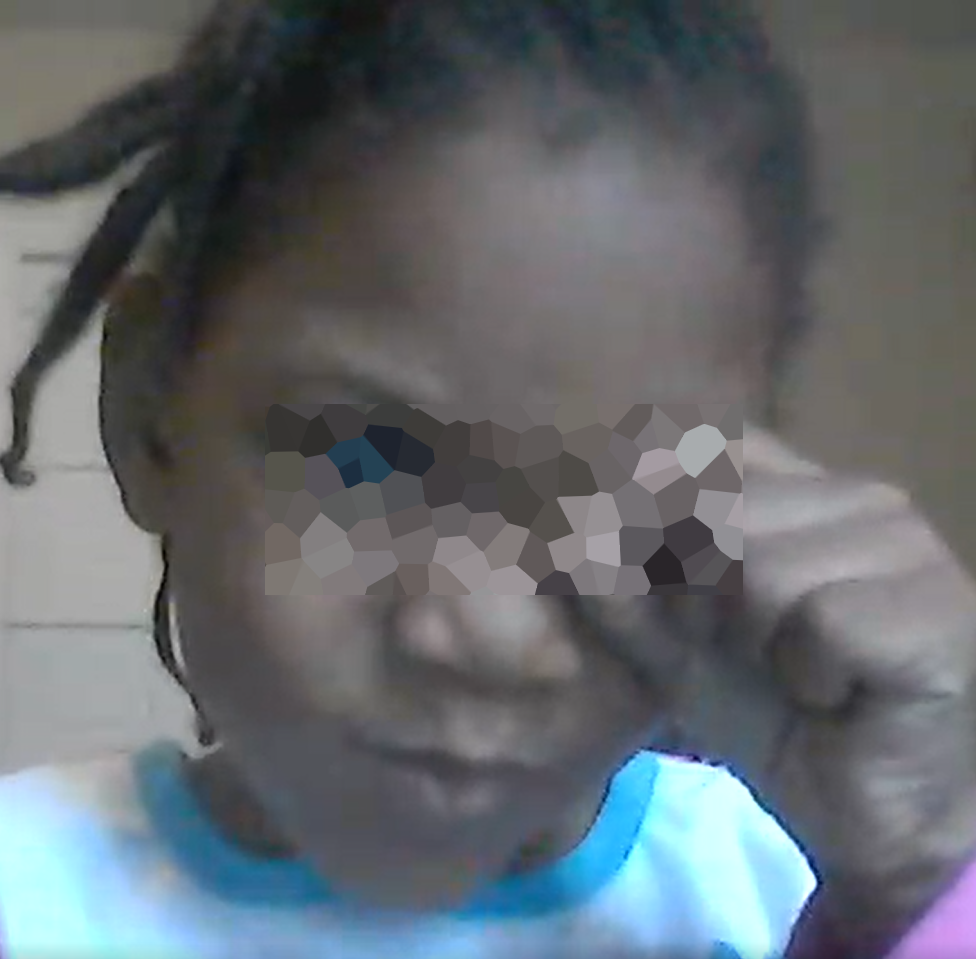}\hfill
    \includegraphics[width=.35\columnwidth]{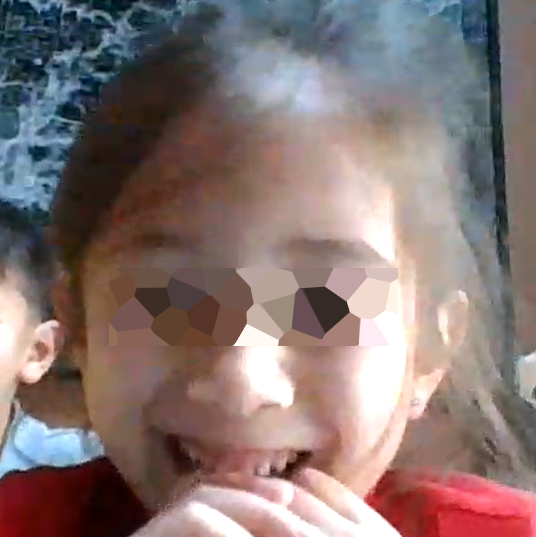}
    \caption{Examples of (from left to right) \textit{eyebrow raise, eyebrow scrunch, hand on face, funny face, and smile}}
    \label{fig:samples}
\end{figure*}
\noindent \textbf{When were children annotated as uncertain?} \\
The annotations are split 13.8/5.3/80.9 between the labels \textit{uncertain}/\textit{unclear}/\textit{non-uncertain}. 
Of all the annotated trials, 79.3\% were correct, of which 82.4\% were rated as having no uncertainty.
In other words, most trials are within the children's capability and confidence. 
While a significant class imbalance exists between uncertain and non-uncertain trials, the distribution is realistic.

\noindent \textbf{Are uncertainty and task difficulty related?} \\
As shown in Figure \ref{fig:distribution}, uncertainty was found to be highly correlated with task difficulty, $r(58) = -.927, p<.01$. Both ratio of dot size and size control factor into the difficulty of a trial (with a smaller ratio and the presence of size control both indicating a harder trial).

\noindent \textbf{Are uncertainty and task performance related?} \\ 
Despite the high correlation between uncertainty and task difficulty, there was no substantial correlation between uncertainty and task correctness, $r(58) = .290, p > .4$.
This makes sense as the accuracy and ease of a task are not necessarily intertwined; a participant may make a mistake on an easy trial or get lucky on a difficult trial.

\noindent \textbf{How do demographics affect uncertainty?} \\
We analyze participant demographics in terms of age and gender. Regarding the frequency of expressing uncertainty, we found that the average participant age of the uncertain trials is younger than that of the non-uncertain trials, though they are not significantly different.
Similarly, female participants have slightly more uncertain trials, and male participants have slightly more non-uncertain trials, but the results are insignificant.

However, we did find gender differences in the types of cues used to express uncertainty.
Female participants exhibited more of the \textit{filled pause} cue, while male participants exhibited more of the \textit{smile} and \textit{shoulder movement} cues. The full table of comparisons can be found in Appendix Table \ref{tab:total_v_uncertain}.

\begin{figure}[!h]
    \centering
    \includegraphics[width=\columnwidth]{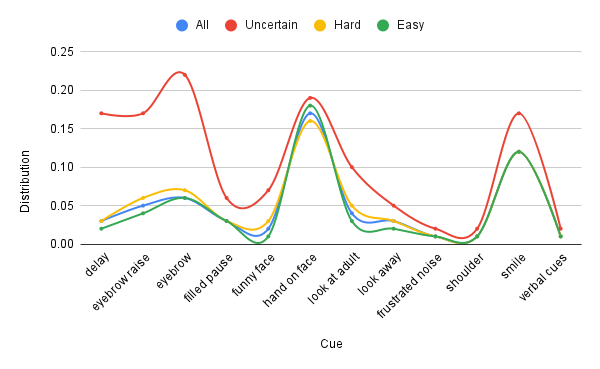}
    \caption{Distribution of uncertainty cues across all/uncertain, difficult/easy. We can see that \textit{delay, eyebrow raise,} and \textit{eyebrow scrunch} are significantly more frequent in uncertain trials.}
    \label{fig:distribution_plot}
\end{figure}

\noindent \textbf{Which cues occur the most?} \\
\label{sec:uncertain_distr}
The percentage in which each cue appears in all uncertain trials can be found in Figure~\ref{fig:distribution_plot}. 
We can see that in general, \textit{hand on face} and \textit{smile} are the most common cues, appearing in 17\% and 12\% of all trials, respectively. 

Notably, during uncertain trials, while \textit{hand on face} and \textit{smile} remain common, other cues also appear more frequently. In particular, \textit{eyebrow scrunch, eyebrow raise} and \textit{delay} are now equally, if not more common, appearing in 22\% and 17\% of all uncertain trials. This is promising, as cues frequently appearing in uncertain trials but not so common throughout all trials can be valuable indicators of uncertainty.

\noindent \textbf{Which cues occur in difficult trials as opposed to easy trials?}\\
The percentage that each cue appears in hard and easy trials can also be found in Figure~ \ref{fig:distribution_plot}. Hard trials are defined as half of the trials with a more difficult ratio (1.11 to 1.17), and easy trials are half with ratios of 1.25 to 2.

We find that if the trial is hard, the participant is slightly more likely to exhibit more of our studied cues overall. In particular, the participant is likelier to exhibit the \textit{look at adult} or \textit{funny face} cues. If the trial is easy, the participant might display \textit{ hand on face} instead.

This shows support for the potential to differentiate between stages of uncertainty. Namely, if a child or student expresses uncertainty at a more manageable task, this could be out of a lack of confidence \textit{(I'm generally familiar with this and have an idea on how to do it, but I need a little help.)} or another factor that may entail minor assistance. Meanwhile, facing a more challenging or perhaps even a completely new task, they may feel a more difficult uncertainty \textit{(I don't know where to start.)} requiring more involved guidance.

This possible distinction in stages of uncertainty may open the door for a more precise intervention in the context of education and tutoring systems. For instance, if notified about a student exhibiting the former uncertainty, the teacher might engage with small hints and encouragement to maximize the student's learning. However, if a student shows the latter uncertainty, the teacher can offer hands-on guidance, such as checking foundational concepts.

%% file: sections/4_computational.tex
\section{Approach}
\label{sec:computational}

With the goal of predicting uncertainty from multimodal signals, we conducted experiments with three approaches: learning uncertainty from proposed cues, a multimodal transformer-based model, and an ensemble learning approach that first predicts cues from the multimodal features and then predicts uncertainty from those cues.


\subsection{Multimodal Features}
We take facial action units, gaze direction, and facial pose from the OpenFace toolkit \cite{8373812} for video features. Action units (AUs) are coded for facial muscle movements, which indicate various facial expressions \cite{tian2001recognizing}. 
For audio features, we extracted glottal source and spectral envelope features using \citet{covarep}(v1.4.2). 
For text features, we then passed GloVe embeddings of each trial's annotated transcription to the model \cite{pennington-etal-2014-glove}. It should be noted that as a task that does not ask the participant to speak, most trials contain no text.

\subsection{MulT Model}
Given the cost of annotation, an ideal uncertainty prediction system would take multimodal data of the participant as features and be able to make real-time predictions on the participant's level of uncertainty.
To test this goal, we first experiment with an end-to-end model. Specifically, we use the Multimodal Transformer proposed in \citet{yaohungt} on audio, video, and text data from videos of the participants. This model is uncertainty cue-agnostic, as it contains no information about our annotated cue categories.

\subsection{Contrastive Learning}
Our dataset has high-dimensional features: 710 dimensionalities for each video frame, 71 for each second of corresponding audio, and 30 for each word in the corresponding text. Training a prediction model end-to-end in a high-dimensional feature space focuses on local differences in the latent space instead of the global relationships between classes. We also tested a contrastive learning procedure with a custom loss function to overcome this challenge. This method encourages the model to learn representations that are close for positive pairs and far apart for negative pairs and better discriminates between different classes. The details of the loss function and our weighted sampling strategy are given in Appendix~\ref{ap:contrastive}.

\subsection {Annotation-based Ensemble learning}
\begin{figure}[h]
    \includegraphics[width=\columnwidth]{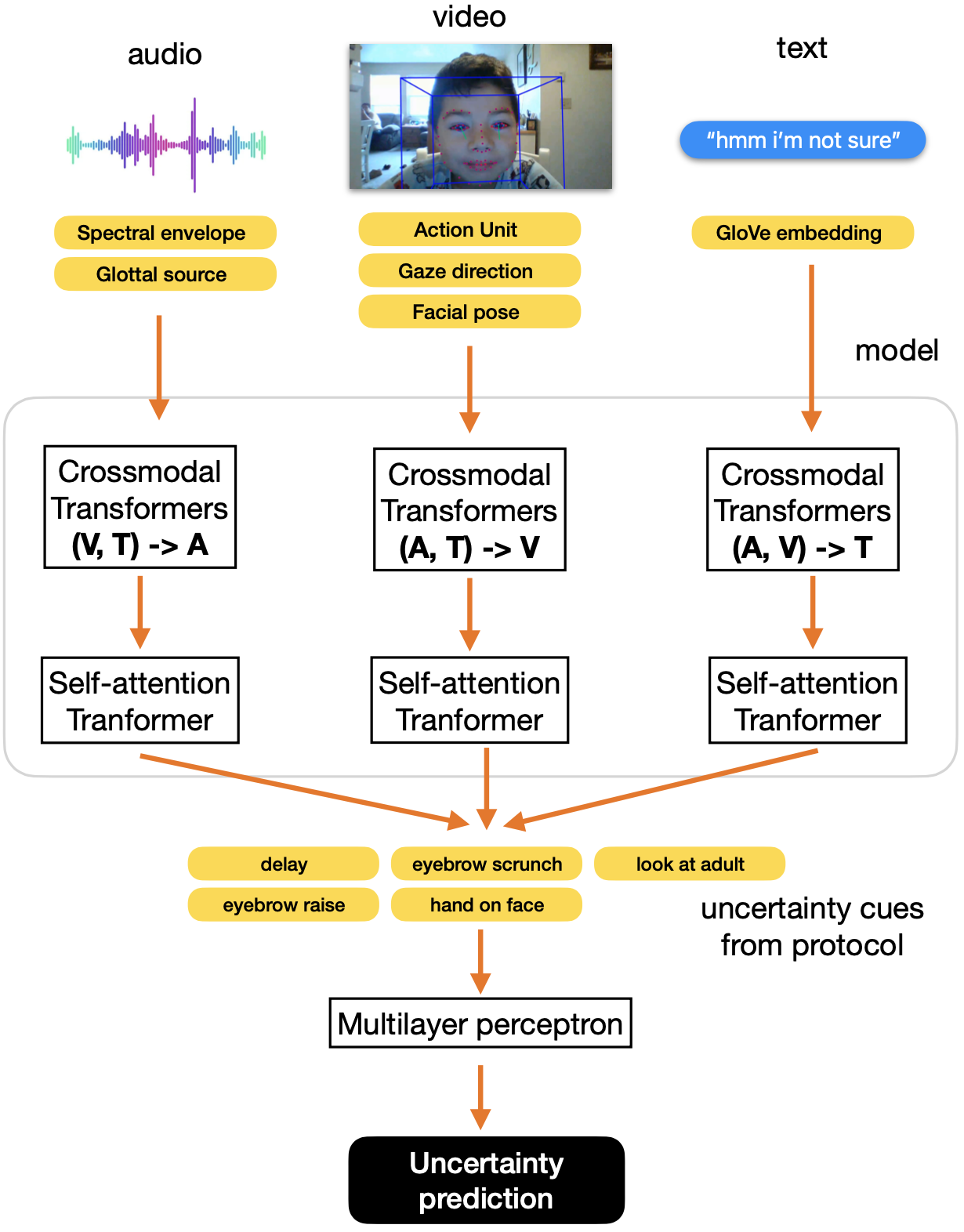}
    \caption{This figure shows the architecture of the ensemble learning model. First, cross-modal transformers learn the attention across the features of each modality with each of the other two modalities' low-level features. Then, using the fused features, self-attention transformers predict the present uncertainty cues, which are then passed into a multilayer perceptron to output the final prediction of whether or not uncertainty is present.}
    \label{fig:architecture}
\end{figure}

We further propose an ensemble learning approach illustrated in Figure \ref{fig:architecture}, that first predicts each of the annotator cues found to be significantly correlated to the annotator prediction of uncertainty and then predicts uncertainty using the trained classifier.
We compare this proposed model to the previous end-to-end multimodal transformer and the unimodal transformers. We choose only to predict the cues that were used in the decision tree classifier (i.e., \textit{delay, eyebrow raise, eyebrow scrunch, look at adult}, and \textit{hand on face}).
\textbf{Multimodal transformer model} is used as the classifier for uncertainty.

%% file: sections/5_results.tex
\section{Experimental Evaluation}
\label{sec:results}
In order to determine the viability of a multimodal uncertainty prediction model, we detail the results from each of our computational experiments and models. The implementation details of the experiments and models are given in the Appendix~\ref{sec:implementation}.  

\begin{table}[h]
    \centering
    \begin{tabular}{cc|rrr}
        \toprule
        & \textbf{Model} & \textbf{F1} & \textbf{MAE} & \textbf{$R^2$} \\
        \midrule
       \multirow{4}{*}{\begin{sideways} No Cue \end{sideways}} & Basic & \textit{.7011} & \textit{.3603} & \textit{-.2419} \\
        & Adult & \textit{.6397} & \textit{.8553} & \textit{-2.7119} \\
        & MulT & .8120 & \textbf{.2307} & \textbf{-.1391}	\\
        & CL + W & \textbf{.8216} & .3076 & -.5412 \\
        \midrule
         \multirow{2}{*}{\begin{sideways} Cue 
        \end{sideways}} & & & \\
        & \multirow{-2}{*}{Ensemble} & \multirow{-2}{*}{\textbf{.8366}} & \multirow{-2}{*}{\textbf{.2222}} & \multirow{-2}{*}{-.1250}\\
        \bottomrule
    \end{tabular}
    \caption{This table shows the results of different models. The "Basic" refers to the MLP baseline, and the ``Adult" refers to the adult uncertainty baseline. ``MulT" refers to the Multimodal Transformer, and ``CL + W" is the MulT model with contrastive learning and weighted sampling. "Ensemble" refers to the cue-based ensemble model. No Cue and Cue indicate whether the model uses the identified cues as intermediate features.}
\label{tab:modal_results_combined}
\end{table}

\textbf{Baselines}
We employed two baseline models for comparison. The first is a simple multimodal neural network that separately processes video, audio, and text inputs and combines the features for a three-class softmax classification. The second is a detection model trained with adult data that takes visual information \cite{adult}. This second baseline is a traditional machine-learning approach using SVMs and LBP descriptors. 

\textbf{Metrics}
In our experiment, we utilized three key metrics to evaluate the performance of our model. The weighted F1 score was employed to account for any class imbalance and provide a more comprehensive assessment of the model's precision and recall. Mean Absolute Error (MAE) was used to measure the average magnitude of the errors in our predictions, illustrating the model's ability to minimize deviations from the actual values. Lastly, the R-square statistic was employed to quantify the proportion of variance in the dependent variable explained by the model, offering insight into the overall goodness of fit and the model's explanatory power.

\subsection{Results}
We report a weighted F1 score of .8216 and a mean absolute error of .3076 on the cue-agnostic end-to-end model with reweighted class labels, as seen in Table \ref{tab:modal_results_combined}. Full results for each modality can also be found in the same table. The cue-aware ensemble model shows improvements in both weighted F1 and MAE over the multimodal transformer model. Contrastive learning and weighted sampling improve the performance but are subpar compared to the cue-based ensemble method. The intermediate prediction of cues like \textit{delay} that are a vital indicator but may be challenging to learn in an end-to-end model may play a role in this performance.

\noindent \textbf{Modality Ablations}
After doing an ablation study on the modalities for the cue-agnostic models, we find that the text and audio modality report the best scores overall, as shown in \ref{tab:result_ablation}. This is unexpected due to participant speech being scarce. However, when participants do talk, they usually express their feelings about the task. For instance, participants may say ``This is easy!" or ``I don't know," tell the adult if the trial is hard or begin counting. As a result, the text modality could be less noisy than the video and audio modalities. We note that the particular task does not request verbal responses from the participants. Thus, we expect that with a task that entices a verbal response, such as question answering, there may be more contribution from the text and audio modalities. 



\begin{table}[!h]
    \centering
    \begin{tabular}{c|r r r}
        \toprule
        \textbf{Model} & \textbf{F1} & \textbf{MAE} & \textbf{$R^2$}  \\
        \midrule
        CL + W & \textbf{.8216} & .3076 & -.5412 \\
        \midrule
        Video only & .7991 & .2820& -.4072 \\
        Text only & \textit{.8056} & \textit{.2564} & \textit{-.2731} \\
        Audio only & \textit{.8056} & \textit{.2564} & \textit{-.2731} \\ 
        \bottomrule
    \end{tabular}
    \caption{F1, MAE, and $R^2$ results for the best performing cue-agnostic model (weighted) with the ablation studies for all the modalities. The model performs the best with all the components, but the most influential modality is text/audio.}
\label{tab:result_ablation}
\end{table}

%% file: sections/6_conclusion.tex
\section{Conventions of Expressing Uncertainty: A Case Study in Different Age Groups}
\label{sec:case_study}
\begin{figure}[!h]
    \centering
        \includegraphics[width=\columnwidth]{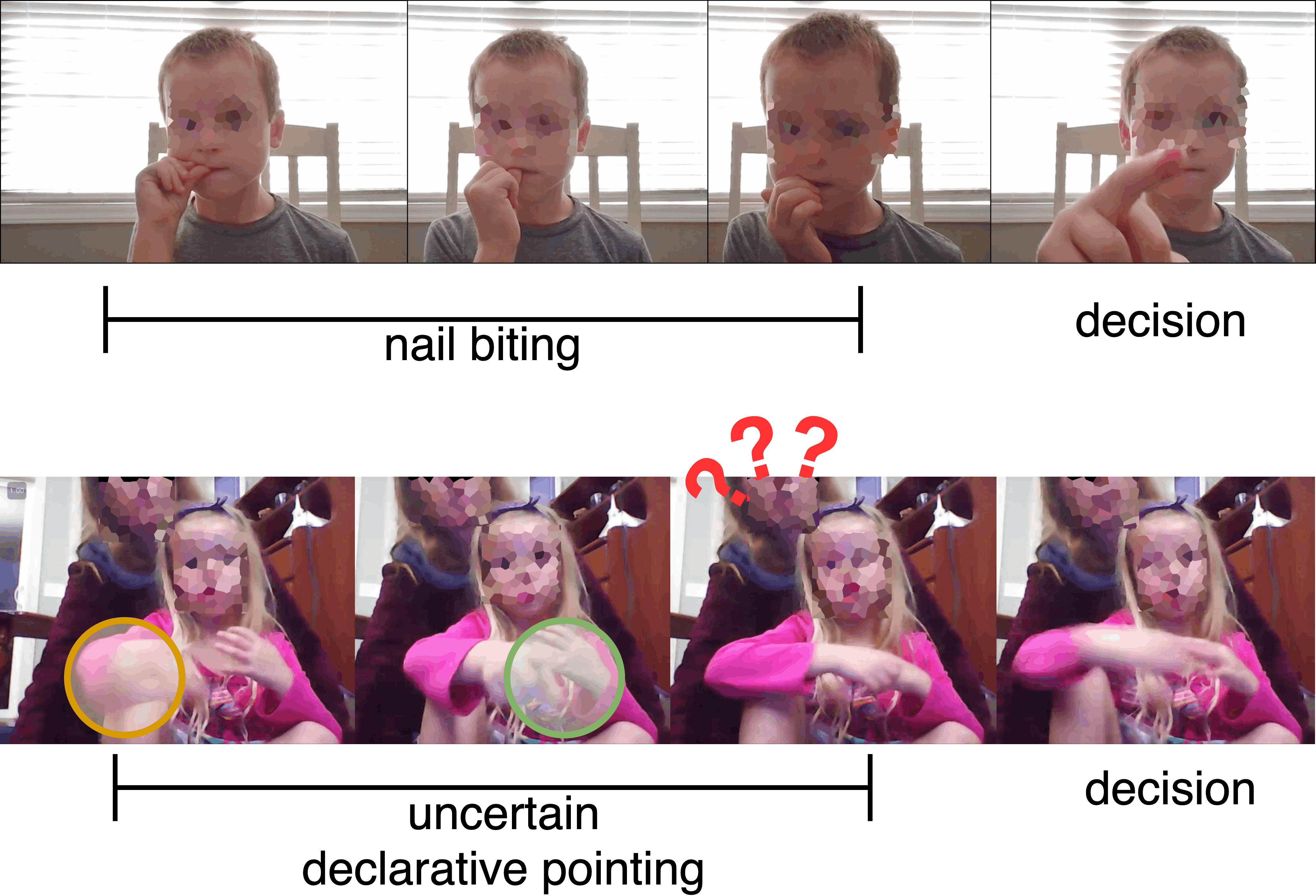}
    \caption{This figure shows two complex behavior patterns by children: nail-biting and uncertain declarative pointing. The top sequence belongs to the oldest male, and the bottom sequence belongs to the youngest female.}
    \label{fig:conventions}
\end{figure}

From our annotated data, we find that for older children ($>2150$ days old), less parental guidance is present, faster decision-making is observed, less diverse facial expressions are present, and more verbal cues are present while expressing uncertainty, which increases the performance of the models for 5-year-olds Table~\ref{tab:result_age_group}. In addition, certain behavior patterns that children of different age groups display are context-dependent, convention-oriented, and personality-specific, making it difficult to identify only through visual and textual modalities. Some of these behaviors that we investigate here are nail-biting, pointing, and social facilitation (see Figure~\ref{fig:conventions}).

\begin{table}[!h]
    \centering
    \resizebox{\columnwidth}{!}{
    \begin{tabular}{c|r r r|r r r}
        \toprule
        & \multicolumn{3}{c|}{\textbf{4 year old}} & \multicolumn{3}{c}{\textbf{5 year old}} \\
        \textbf{Model} & \textbf{F1} & \textbf{MAE} & \textbf{$R^2$} & \textbf{F1} & \textbf{MAE} & \textbf{$R^2$} \\
        \midrule
        Basic & \textit{.69} & \textit{.32} & \textit{-.24} & \textit{.73} & \textit{.33} & \textit{-.21}\\
        Adult & \textit{.57} & \textit{1.06} & \textit{-4.40}  & \textit{.62} & \textit{1.0} & \textit{-8.13}\\
        MulT & .78 & .26 & -.39 & .81 & .26 & -.27\\
        CL + W & \textbf{.79} & \textbf{.22} & \textbf{-.15} & \textbf{.81} & \textbf{.23} & \textbf{-.14}\\
        \bottomrule
    \end{tabular}
     }
    \caption{This table shows the results of the models between different age groups. There are slight inference differences between the 4 and 5-year-old groups. These performance changes are dependent on the conventions of uncertainty and age.}
\label{tab:result_age_group}
\end{table}

In the developmental psychology literature, nail-biting is either found to be an acquired habit or related to states of nervousness \cite{gilleard1988nailbiting, silber1992treating, ghanizadeh2008association, mcclanahan1995operant, wells1998severe}. This behavior is hard to classify as stress-induced or uncertainty-induced. Hence, a context-dependent analysis of the person using skeletal features can help decide.

Pointing (see Figure~\ref{fig:conventions}) is another context-dependent occurrence. When the child is uncertain, the pointing to options also becomes ambiguous, and the parent needs to ask a follow-up grounding clarification question, such as "Which one?". This behavior pattern involves spatial placement of options and understanding the boundaries between them. Younger children prefer ambiguous pointing gestures to conventional and visible cues of uncertainty. This type of declarative pointing is observed to be a way of engaging with the parent, pointing to a theory of mind (ToM) understanding by the children \cite{pointing}. Skeletal and ToM modeling can help make prediction performance better.

Another behavior pattern is social facilitation. Younger children prefer to be together with their parents while solving tasks. Older children follow verbal conventions and reduce the vividness of their facial expressions, while younger children exaggerate their facial expressions and rely more on social facilitation factors. Similar behavior patterns happen in adults in a competitive atmosphere where social facilitation has different effects on an individual's facial expressions \cite{buck1992social, katembu2022effects}. ToM and multi-party dialogue modeling can increase the performance of uncertainty understanding models.

Uncertainty is context-dependent -- some children are naturally more fidgety or shy. So predicting uncertainty on an isolated trial basis may lead to less accurate results. As a result, one interesting question is how to incorporate contextual features about the participant's personality and recent cognitive states to make more informed predictions.

\section{Conclusion}
\label{sec:conclusion}
In this paper, we explored the task of predicting uncertainty in young children from an annotated dataset that we introduced with a multimodal transformer-based model. We discover that demographic and trial difficulty can affect the frequency of certain cues. Moreover, trial difficulty strongly correlates with uncertainty, but trial performance interestingly does not. There is still room for improvement in task performance by transformer models, which means that more data or more complicated task setups are needed to study uncertainty properly. Our dataset--which we make available for research purposes--and protocol provide future researchers with additional tools to predict uncertainty using multimodal cues to facilitate human-human and human-AI dialogue.

\section{Ethics}
\label{sec:ethics}
Due to the sensitive nature of the video data of children and their privacy, we are only making some portion of the data publicly available with the consent of the parents of the children. All the images used in this paper are from the videos that are from children and parents who have given consent to share their video data publicly.

%% file: sections/appendix.tex
\section{Percentages of each cue}
\begin{figure}[!h]
    \centering
    \includegraphics[width=\columnwidth]{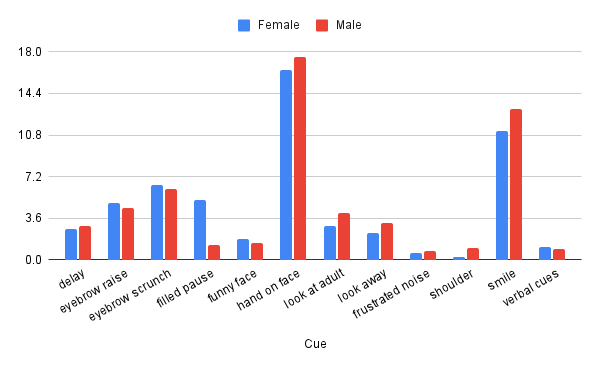}
    \caption{Female and male participants show small differences in frequency of displaying certain cues.}
    \label{fig:male_female}
\end{figure}

Here we present in Table~\ref{tab:total_v_uncertain}, all the distribution of the uncertainty cues we found in the dataset. We also present more statistics between female and male participants in Figure~\ref{fig:male_female}.

\begin{table*}[t]
    \centering
    \begin{tabular}{|l|rr|rr|rr|}
        \toprule
        \textbf{Cue} & \textbf{All Pct.} & \textbf{Uncertain Pct.} & \textbf{Hard Pct.} & \textbf{Easy Pct.} & \textbf{Female Pct.} & \textbf{Male Pct.}\\
        \midrule
        delay & 0.03 & \textbf{0.17} & 0.03 & 0.02  &  2.7  & 2.93\\
        eyebrow raise & 0.05 & \textbf{0.17}  & \textbf{0.06} & 0.04 & 4.94  & 4.48 \\
        eyebrow scrunch & 0.06 & \textbf{0.22} & 0.07 & 0.06  & 6.49  & 6.15\\
        filled pause & 0.03 & 0.06 & 0.03 & 0.03  & \textbf{5.17 } & 1.26\\
        funny face & 0.02 & 0.07  & \textbf{0.03} & 0.01 &  1.84  &  1.49\\
        hand on face & \textbf{0.17} & \textbf{0.19} & \textbf{0.16} & \textbf{0.18} & 16.44  & 17.53\\
        look at adult & 0.04 & 0.1 & \textbf{0.05} & 0.03 &  2.93  & 4.02\\
        look away & 0.03 & 0.05 & 0.03 & 0.02 & 2.36  & 3.16\\
        frustrated noise & 0.01 & 0.02  & 0.01 & 0.01  & 0.57  & 0.8\\
        shoulder & 0.01 & 0.02 & 0.01 & 0.01 & 0.29  & \textbf{1.03}\\
        smile & \textbf{0.12} & \textbf{0.17} & \textbf{0.12} & \textbf{0.12} & 11.15  & \textbf{13.05} \\
        verbal cues & 0.01 & 0.02 & 0.01 & 0.01& 1.09  & 0.92\\
        \bottomrule
    \end{tabular}
    \caption{Distribution of uncertainty cues across all/uncertain, difficult/easy, and female/male trials. We can see that \textit{delay, eyebrow raise,} and \textit{eyebrow scrunch} are significantly more frequent in uncertain trials. Meanwhile, if the trial is hard, participants are likelier to look at an adult or make a funny face. If the trial is easy, the participant may display \textit{hand on face} instead. Female and male participants also show mild differences in the frequency of displaying certain cues.}
    \label{tab:total_v_uncertain}
\end{table*}

\section{Contrastive Learning Details}
\subsection{Problem Statement}
In this study, we aim to predict the uncertainty of a child based on multimodal inputs, including video, transcripts, and audio. Given the dataset of instances, each containing video (V), transcripts (T), and audio (A) data, our goal is to develop a model that can accurately predict whether a child is uncertain, unclear, or not uncertain. We represent this problem as a function F that maps the input features (V, T, A) to the binary output variable $y \in \{0, 0.5, 1\}$, where 0 denotes not uncertain, 0.5 denotes unclear, and 1 denotes uncertain:

To achieve this, we design a multimodal transformer model that leverages the complementary information in the video, transcripts, and audio data to make predictions. The model is trained on a dataset of labeled examples $(V_i, T_i, A_i, y_i)$, where $i \in \{1, ..., N\}$ and N is the total number of instances. Our objective is to minimize the cross-entropy loss. By minimizing this loss, our model will learn to predict a child's uncertainty level accurately based on the provided multimodal inputs.

\label{ap:contrastive}
The specific contrastive learning loss function, $L(X_1, X_2, L_1, L_2)$, that we are focusing on here is defined as the following:
\begin{equation}
\ L(X,L) = \frac{1}{N_1 * N_2} \sum_{i} \sum_{j} W_{ij} * (m - S_{ij})^2_+
\end{equation}
This function captures the relationship between pairs of data points $X1$ and $X2$, with associated labels $L1$ and $L2$. The cosine similarity, $S_{ij}$, is used to measure the similarity between the data points, and the weighing factor, $W_{ij}$, is used to differentiate between positive and negative pairs. The weighting factor is determined using the Kronecker delta function, $\delta(L1_i, L2_j)$. The $e^{\alpha * \delta(L1_i, L2_j)}$ coefficient ensures that the positive pairs have a greater influence on the learning process where $m$ is a threshold to separate positive and negative pairs and $\alpha$, is a scaling factor. Lastly, $(x)_+$ ensures that only the non-negative values of $x$ are considered.

To further improve our model's performance, we employed a weighted sampling method using the class frequencies' inverse square root. Given a dataset with classes $0$ being certain, $0.5$ being unclear, and $1$ being uncertain, we calculate the weights for each class sample as follows:
\begin{equation}
\ w_i = \frac{1}{\sqrt{\mathrm{N_i}}}, \quad \mathrm{where } i \in \{0, 0.5, 1\}.
\end{equation}

In our case, this weighting scheme assigns higher weights to underrepresented classes, the class of 0.5 (unclear) and 1 (uncertain), which helps balance class sampling probabilities. The inverse square root function is particularly useful as it provides a smooth topology less sensitive to small changes in class frequencies than other weighing functions.

\section{Implementation Details}
\label{sec:implementation}
\subsection{Experimental setup}
The argmax of the label probabilities was taken as the output layer.
All networks were trained for 40 epochs with a batch size of 24. Both raw and weighted cross entropy loss were used to train two versions of the model. Class weights were set based on the distribution of train set samples to mitigate class imbalance issues.

We employ a 75-10-15 training-dev-test split. For each result, we report the average across three different seeds. We also run the model on every single modality to provide unimodal baselines.

Additionally, we have age information for each participant, so we divided the dataset into two age groups: 4-year-olds and 5-year-olds. This division will allow us to investigate potential differences between the two age groups as shown in the 4-year-old and 5-year-old tabs in Table \ref{tab:modal_results_combined}.

\textbf{Transformer Model Details.} 
Our multimodal transformer model is based on a modified version of the Transformer architecture. It consists of five layers, each equipped with five attention heads to capture various contextual relationships within the input data. We used the Stochastic Gradient Descent (SGD) optimizer with an initial learning rate of 0.001 to train our model. To enhance convergence and overall performance, we employed the ReduceLROnPlateau learning rate scheduler, which adjusts the learning rate when the validation loss ceases to improve. We set the reduction factor to 0.1 and patience of 5 epochs for monitoring improvements. Our model was trained on an NVIDIA RTX 4090 GPU, using a batch size of 1. We trained the model for 100 epochs. For models with contrastive learning, we trained the model with additional 10 epochs before real training.